\documentclass[10pt,letterpaper]{article}

\usepackage{cogsci}
\usepackage{pslatex}
\usepackage{apacite}

\usepackage{amsmath, amsthm, amssymb}
\usepackage{graphicx}
\usepackage{color}
\usepackage{booktabs}
\usepackage{lipsum}
\usepackage{xcolor}
\usepackage{subcaption}

\title{Relational inductive bias for physical construction in humans and machines}

\author{\large \bf Jessica B.~Hamrick$^{*,1}$ (jhamrick@google.com), Kelsey R.~Allen$^{*,2}$ (krallen@mit.edu), \\
  \large \bf Victor Bapst$^1$ (vbapst@google.com), Tina Zhu$^1$ (tinazhu@google.com), \\
  \large \bf Kevin R.~McKee$^1$ (kevinrmckee@google.com), Joshua B.~Tenenbaum$^2$ (jbt@mit.edu), \\
  \large \bf Peter W.~Battaglia$^1$ (peterbattaglia@google.com) \\
  $^1$DeepMind; London, UK\\
  $^2$Department of Brain and Cognitive Sciences; MIT; Cambridge, MA, USA}

\newcommand{\AvgHumanReward}[0]{900}
\newcommand{\MinHumanReward}[0]{468}
\newcommand{\MaxHumanReward}[0]{1154}

\newcommand{\TrialRewardCorr}[0]{$r=0.15\mathrm{,\ 95\%\ CI\ }[-0.01, 0.30]$}

\newcommand{\FullVsPartialSevenBlocks}[0]{$M=0.01\mathrm{,\ 95\%\ CI\ }[-0.03, 0.05]$}
\newcommand{\FullVsPartialTenBlocks}[0]{$M=0.05\mathrm{,\ 95\%\ CI\ }[-0.01, 0.10]$}
\newcommand{\FullVsPartialSevenBlocksFC}[0]{$M=-0.04\mathrm{,\ 95\%\ CI\ }[-0.08, 0.00]$}
\newcommand{\FullVsPartialTenBlocksFC}[0]{$M=0.44\mathrm{,\ 95\%\ CI\ }[0.27, 0.61]$}
\newcommand{\PaddedMLPReward}[0]{$M=78.00\mathrm{,\ 95\%\ CI\ }[-140.00, 296.00]$}
\newcommand{\HumanTopToBottomSlopes}[0]{$\beta=-0.07\mathrm{,\ 95\%\ CI\ }[-0.08, -0.06]$}

\newcommand{\HumanProportionFP}[0]{73\%}
\newcommand{\HumanProportionFPTest}[0]{$N=3901$, $p < 0.001$}
\newcommand{\ModelProportionFP}[0]{41\%}
\newcommand{\ModelProportionFPTest}[0]{$N=155$, $p < 0.05$}

\newcommand{\HumanPctInvalid}[0]{1.3\% (out of $N=6454$)}
\newcommand{\MLPPctInvalid}[0]{46\% (out of $N=417$)}

\newcommand{\GNFCPctInvalid}[0]{31\% (out of $N=1345$)}

\newcommand{\HumanPctRepeat}[0]{0.29\% out of $N=6454$}

\newcommand{\GNFCVsMLPReward}[0]{$M=883.60\mathrm{,\ 95\%\ CI\ }[719.40, 1041.00]$}
\newcommand{\GNVsGNFCReward}[0]{$M=183.20\mathrm{,\ 95\%\ CI\ }[73.20, 302.40]$}

\newcommand{\SimulationVsGNSReward}[0]{$M=156.20\mathrm{,\ 95\%\ CI\ }[70.80, 249.60]$}
\newcommand{\FirstActionRT}[0]{$t=4.43\mathrm{s}\mathrm{,\ 95\%\ CI\ }[4.30\mathrm{s}, 4.56\mathrm{s}]$}
\newcommand{\OtherActionsRT}[0]{$t=2.07\mathrm{s}\mathrm{,\ 95\%\ CI\ }[2.00\mathrm{s}, 2.15\mathrm{s}]$}
\newcommand{\FirstSecondBlockRTDiff}[0]{$t=4.48\mathrm{s}\mathrm{,\ 95\%\ CI\ }[4.34\mathrm{s}, 4.62\mathrm{s}]$}
\newcommand{\RTFirstGlueAnova}[0]{$F(1, 12878)=149.14$, $p < 0.001$}
\newcommand{\RTNumBlocksAnova}[0]{$F(1, 12878)=429.68$, $p < 0.001$}
\newcommand{\RTInteractionAnova}[0]{$F(1, 12878)=14.57$, $p < 0.001$}

\newcommand\blfootnote[1]{%
  \begingroup
  \renewcommand\thefootnote{}\footnote{#1}%
  \addtocounter{footnote}{-1}%
  \endgroup
}

\begin{document}

\maketitle\blfootnote{$^*$Denotes equal contribution.}

\begin{abstract}

While current deep learning systems excel at tasks such as object classification, language processing, and gameplay, few can construct or modify a complex system such as a tower of blocks.
We hypothesize that what these systems lack is a ``relational inductive bias'': a capacity for reasoning about inter-object relations and making choices over a structured description of a scene.
To test this hypothesis, we focus on a task that involves gluing pairs of blocks together to stabilize a tower, and quantify how well humans perform.
We then introduce a deep reinforcement learning agent which uses object- and relation-centric scene and policy representations and apply it to the task.
Our results show that these structured representations allow the agent to outperform both humans and more na\"ive approaches, suggesting that relational inductive bias is an important component in solving structured reasoning problems and for building more intelligent, flexible machines.

\textbf{Keywords:} 
physical construction; reinforcement learning; deep learning; relational reasoning; object-based reasoning
\end{abstract}

\section{Introduction}

Human physical reasoning---and cognition more broadly---is rooted in a rich system of knowledge about objects and relations \shortcite{Spelke2007} which can be composed to support powerful forms of combinatorial generalization. Analogous to von Humboldt's characterization of the productivity of language as making ``infinite use of finite means'', objects and relations are the building blocks which help explain how our everyday scene understanding can operate over infinite scenarios.
Similarly, people \emph{interact} with everyday scenes by leveraging these same representations. 
Some of the most striking human behavior is our ubiquitous drive to build things, a capacity for composing objects and parts under relational constraints, which gives rise to our most remarkable achievements, from the pyramids to space stations.

One of the fundamental aims of artificial intelligence (AI) is to be able to interact with the world as robustly and flexibly as people do.
We hypothesize that this flexibility is, in part, afforded by what we call \emph{relational inductive bias}.
An inductive bias more generally is the set of assumptions of a learning algorithm that leads it to choose one hypothesis over another independent of the observed data.
Such assumptions may be encoded in the prior of a Bayesian model \shortcite{Griffiths2010}, or instantiated via architectural assumptions in a neural network.
For example, the weight-sharing architecture of a convolutional neural network induces an inductive bias of translational invariance---one we might call a ``spatial inductive bias'' because it builds in specific assumptions about the spatial structure of the world.
Similarly, a relational inductive bias builds in specific assumptions about the \emph{relational} structure of the world.

While logical and probabilistic models naturally contain strong relational inductive biases as a result of propositional and/or causal representations, current state-of-the-art deep reinforcement learning (deep RL) systems  rarely use such explicit notions and, as a result, often struggle when faced with structured, combinatorial problems.
Consider the ``gluing task'' in Figure~\ref{fig:task}, which requires gluing pairs of blocks together to cause an otherwise unstable tower to be stable under gravity.
Though seemingly simple, this task is not trivial. It requires (1) reasoning about variable numbers and configurations of objects; (2) choosing from variably sized action spaces (depending on which blocks are in contact); and (3) selecting where to apply glue, from a combinatorial number of possibilities.
Although this task is fundamentally about physical reasoning, we will show that the most important type of inductive bias for solving it is relational, not physical: the physical knowledge can be learned, but relational knowledge is much more difficult to come by.

We instantiate a relational inductive bias in a deep RL agent via a ``graph network'', a neural network for relational reasoning whose relatives \shortcite{Scarselli2009} have proven effective in theoretical computer science \shortcite{Dai2017}, quantum chemistry \shortcite{gilmer2017neural}, robotic control \shortcite{Wang2018}, and learning physical simulation \shortcite{Battaglia2016,Chang2017}.
Our approach contrasts with standard deep learning approaches to physical reasoning, which are often computed holistically over a fixed representation and do not explicitly have a notion of objects or relations \shortcite<e.g.>{Lerer2016,Li2016}.
Further, our work focuses on interaction, while much of the work on physical reasoning has focused on the task of prediction \shortcite<e.g.>{Fragkiadaki2016,Mottaghi2016a,Mottaghi2016b,stewart2017label,Bhattacharyya2018} or inference \shortcite<e.g.>{Wu2016,Denil2017}.
Perhaps the most related works to ours are \shortciteA{Li2017} and \shortciteA{Yildirim2017}, which both focus on building towers of blocks. However, while \shortciteA{Li2017}'s approach is learning-based, it does not include a relational inductive bias; similarly, \shortciteA{Yildirim2017}'s approach has a relational inductive bias, but no learning.

This goal of this paper is not to present a precise computational model of how humans solve the gluing task, nor is it to claim state-of-the-art performance on the gluing task.
Rather, the goal is to characterize the type of inductive bias that is necessary in general for solving such physical construction tasks.
Our work builds on both the broader cognitive literature on relational reasoning using graphs \shortcite<e.g.>{Collins1975,Shepard1980,Griffiths2007,Kemp2008} as well as classic approaches like relational reinforcement learning \shortcite{Dvzeroski2001}, and represents a step forward by showing how relational knowledge can be disentangled from physical knowledge through relational policies approximated by deep neural networks.

The contributions of this work are to: (1) introduce the gluing task, an interactive physical construction problem that requires making decisions about relations among objects;
(2) measure human performance in the gluing task;
(3) develop a deep RL agent with an object- and relation-centric scene representation and action policy;
and (4) demonstrate the importance of relational inductive bias by comparing the performance of our agent with several alternatives, as well as humans, on both the gluing task and several control tasks that isolate different aspects of the full problem.

\section{The Gluing Task}

\subsubsection{Participants}

We recruited 27 volunteers from within DeepMind.
Each participant was treated in accordance with protocols of the UCL Research Ethics Committee, and completed 144 trials over a one-hour session.
Two participants did not complete the task within the allotted time and were excluded from analysis, leaving 25 participants total.

\subsubsection{Stimuli and Design}

The stimuli were towers of blocks similar to those used by \shortciteA{Battaglia2013} and \shortciteA{Hamrick2016}.
Towers were created by randomly placing blocks on top of each other, with the following constraints: the tower was constructed in a 2D plane, and each block except the first was stacked on another block.
The set of towers was filtered to include only those in which at least one block moved when gravity was applied.
In an initial \emph{practice} session, nine unique towers (1 each of 2-10 blocks) were presented in increasing order of size.
In the \emph{experimental} session, 135 unique towers (15 each of 2-10 blocks), which were disjoint from the practice set, were presented in a random order in 5 sets of 27.
Participants earned points depending on how well they performed the gluing task.
They lost one point for each pair of objects they tried to glue, and earned one point for each block that remained unmoved after gravity was applied.
As a bonus, if participants used the minimum amount of glue necessary to keep the tower stable, they received 10 additional points.
The maximum possible scores in the practice and experimental sessions were 131 points and 1977 points, respectively.

\subsubsection{Procedure}

Each trial consisted of two phases: the \emph{gluing} phase, and the \emph{gravity} phase.
The trial began in the gluing phase, during which a static tower was displayed on the screen for an indefinite amount of time.
Participants could click on one object (either a block or the floor) to select it, and then another object to ``glue'' the two together.
Glue was only applied if the two objects were in contact.
If glue had already been applied between the two objects, then the glue was removed.
Both these actions---applying glue to non-adjacent objects and ungluing an already-glued connection---still cost one point.\footnote{While this choice of reward structure is perhaps unfair to humans, it provided a fairer comparison to our agents who would otherwise not be incentivized to complete the task quickly.}
To finish the gluing phase, participants pressed the ``enter'' key which triggered the gravity phase, during which gravity was applied for 2s so participants could see which blocks moved from their starting positions.
Finally, participants were told how many points they earned and could then press ``space'' to begin the next trial.
Physics was simulated using the Mujoco physics engine \shortcite{Todorov2012} with a timestep of 0.01.
After the experiment was completed, participants completed a short survey.

\begin{figure}[t!]
\begin{center}
\includegraphics[width=0.4\textwidth]{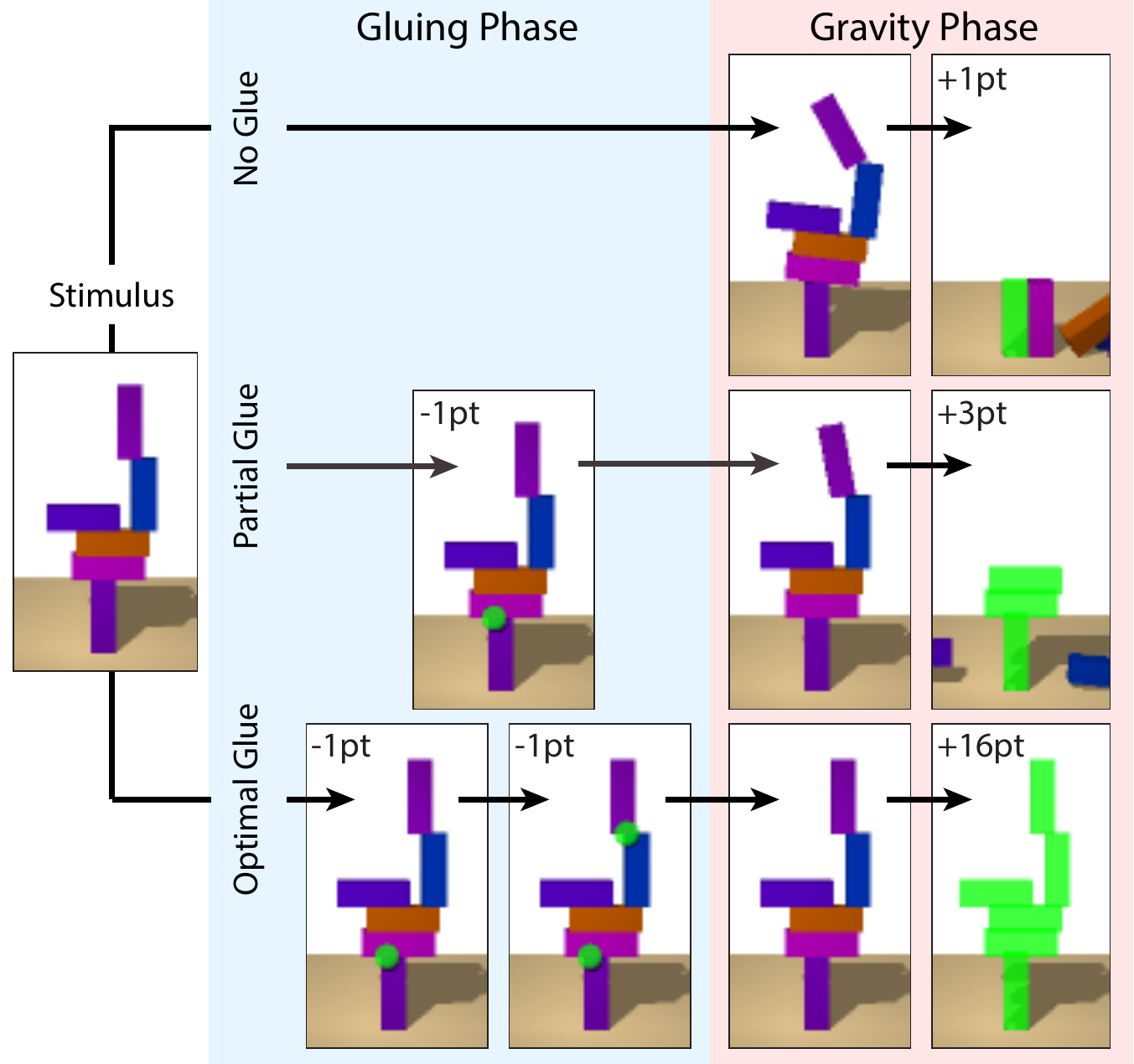}
\caption{\textbf{The gluing task.} Given an unstable tower of blocks, the task is to glue pairs of blocks together to keep the tower stable. Three examples of performing the task are shown here. Green blocks in the gravity phase indicate stable blocks. Top: no glue is used, and only one block remains standing (+1 points). Middle row: one glue is used (-1 points), resulting in three blocks standing (+3 points). Bottom row: two glues are used (-2 points), resulting in a stable tower (+6 points); this is the minimal amount of glue to keep the tower stable (+10 points).
See \url{https://goo.gl/f7Ecw8} for a video demonstrating the task.
}
\label{fig:task}
\end{center}
\end{figure}

\subsubsection{Results}

The gluing task was challenging for the human participants, but they still performed far above chance. We discovered several trends in people's behavior, such as working from top-to-bottom and spending more time before applying the first glue than before subsequent glue.
The results here represent a preliminary exploration of people's behavior in construction tasks, opening the door for future research and providing a baseline comparison for artificial agents.

Participants achieved an average score of \AvgHumanReward{} points, with the lowest score being \MinHumanReward{} points and the highest score being \MaxHumanReward{} points (out of 1977).
There was a small (though not quite significant) effect of learning, with a Pearson correlation of \TrialRewardCorr{} between trial number and average scaled reward (confidence intervals were computed around the median using 10,000 bootstrap samples with replacement; ``scaled rewards'' were computed by normalizing rewards such that 0 corresponded to the reward obtained if no actions were taken, and 1 corresponded to the maximum achievable reward).

Participants' response times revealed they were significantly slower to click on the first block in a pair than the second block, with a difference of \FirstSecondBlockRTDiff{}.
This suggests they had decided on which pair to glue before clicking the first block.
We found that people were significantly slower to choose the first gluing action (\FirstActionRT{}; averages computed using the mean of log RTs) than any subsequent gluing action (\OtherActionsRT{}; \RTFirstGlueAnova{}).
Also, we found an effect of the number of blocks on response time (\RTNumBlocksAnova{}) as well as an interaction between the number of blocks and whether the action was the first glue or not (\RTInteractionAnova{}), with the first action requiring more time \emph{per block} than subsequent actions.
These results suggest that people may either decide where to place glue before acting, or at least engage in an expensive encoding operation of a useful representation of the stimulus.

On an open-ended strategy question in the post-experiment survey, 10 of 25 participants reported making glue selections top-to-bottom, and another 3 reported sometimes working top-to-bottom and sometimes bottom-to-top.
We corroborated this quantitatively by, for each trial, fitting a line between the action number and the height of the glue location, and find their slopes were generally negative (\HumanTopToBottomSlopes{}).

We compared people's choice of glue configuration to optimal glue configurations, and found that people were significantly more likely to apply glue when it was not necessary (\HumanProportionFP{} of errors) than to fail to apply glue when it was necessary (\HumanProportionFPTest{}).
Additionally, participants were very good at avoiding invalid actions: although they had the option for gluing together pairs of blocks that were not in contact, they only did so \HumanPctInvalid{} of the time.
Similarly, participants did not frequently utilize the option to un-glue blocks (\HumanPctRepeat{}), likely because it incurred a penalty.
It is possible that performance would increase if participants were allowed to un-glue blocks without a penalty, enabling them to temporarily use glue as a working memory aid; we leave this as a question for future research.

\begin{figure}[!t]
\begin{center}
\includegraphics[width=0.45\textwidth]{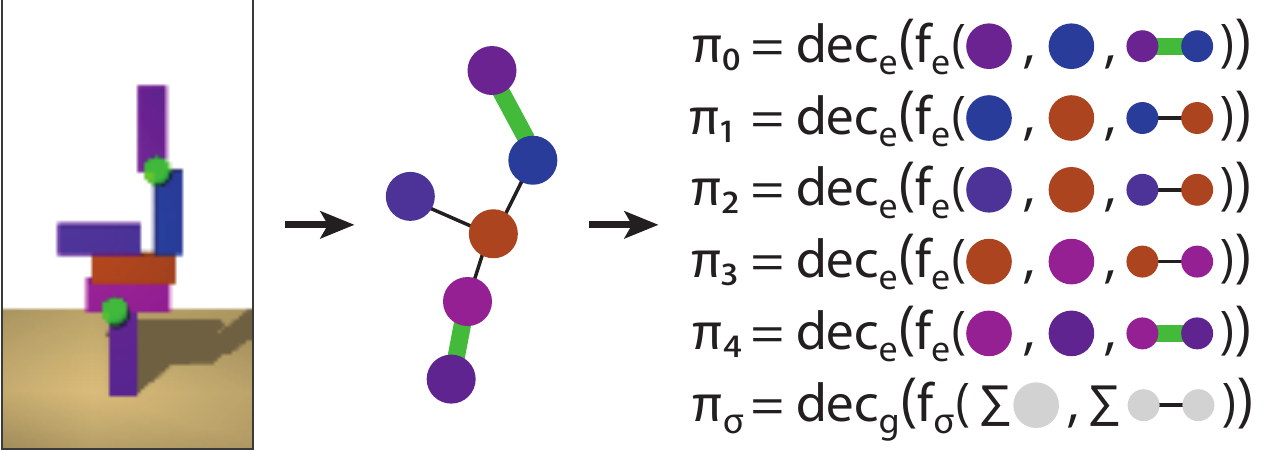}
\caption{\textbf{Graph network agent.} First, the positions and orientations of the blocks are encoded as nodes, and the presence of glue is encoded as edges. These representations are then used to compute a Q-value for each edge, as well as a Q-value for taking the ``stop'' action. See text for details.}
\label{fig:graphnet}
\end{center}
\end{figure}

\section{Leveraging Relational Representations}

What type of knowledge is necessary for solving the gluing task?
Physical knowledge is clearly important, but even that implicitly includes a more foundational type of knowledge: that of objects and relations.
Inspired by evidence that objects and relations are a core part of human cognition \shortcite<e.g.>{Spelke2007}, we focus on decomposing the task into a relational reasoning problem which involves computations over pairs of elements and their relations.

\subsection{Graph Networks}

A key feature of our deep RL agent is that it expresses its decision-making policy as a function over an object- and relation-centric state representation, which reflects a strong relational inductive bias.
Specifically, inside the agent is a \emph{graph network} (GN), a neural network model which can be trained to approximate functions on graphs. A GN is a generalization of recent neural network approaches for learning physics engines \shortcite{Battaglia2016,Chang2017}, as well as message-passing neural networks \shortcite{gilmer2017neural,Scarselli2009}.
GNs have been shown to be effective at solving classic combinatorial optimization problems \shortcite{Dai2017}, inspiring our agent architecture for performing physical construction tasks.

Here, we define a graph as a set of $N$ nodes, $E$ edges, and a global feature $G$.
In the gluing task's ``tower graph'', nodes correspond to blocks; edges correspond to pairs of blocks; and global properties could correspond to any global piece of information, such as the overall stability of the tower. 
A GN takes as input a tower graph, and returns a graph with the same size and shape.
The representation of the nodes, edges, and globals encode semantic information: the node representation corresponds to position ($x$) and orientation ($q$), and the edges to the presence of glue ($u$).
The global features correspond to (or are a function of) the whole graph; for example, this could be the stability of the tower.

Our model architectures first encode the block properties into a distributed node representation $\mathbf{n}_i$ using an encoder, i.e. $\mathbf{n}_i=\mathrm{enc}_n(x_i, q_i; \theta_{\mathrm{enc}_n})$.
For an edge $\mathbf{e}_{ij}$, we similarly encode the edge properties into a distributed representation using a different encoder, i.e. $\mathbf{e}_{ij}=\mathrm{enc}_e(u_{ij}; \theta_{\mathrm{enc}_e})$.
Initially, the global properties are empty and set to zero, i.e. $\mathbf{g}=0$.
With these node, edge, and global representations, the standard GN computes functions over pairs of nodes (e.g., to determine whether those nodes are in contact)\footnote{These functions are learned and thus these examples are not \emph{literally} what the agent is computing, but we provide them here to give an intuition for how GNs behave.}, edges (e.g. to determine the force acting on a block), and globals (e.g. to compute overall stability).
Specifically, the edge model is computed as: $
\mathbf{e}_{ij}^\prime = f_e(\mathbf{n}_i, \mathbf{n}_j, \mathbf{e}_{ij}, \mathbf{g}; \theta_{f_e})$; the node model as $\mathbf{n}_i^\prime = f_n(\mathbf{n}_i, \sum_j \mathbf{e}_{ij}^\prime, \mathbf{g}; \theta_{f_n})$; and the globals model as $\mathbf{g}^\prime = f_g(\mathbf{g}, \sum_i \mathbf{n}_i^\prime, \sum_{i,j} \mathbf{e}_{ij}^\prime; \theta_{f_g})$.
The GN can be applied multiple times, recurrently, where $\mathbf{e}_{ij}^\prime$, $\mathbf{n}_i^\prime$, and $\mathbf{g}^\prime$ are fed in as the new $\mathbf{e}_{ij}$, $\mathbf{n}_i$, and $\mathbf{g}$ on the next step.

Applying the GN to compute interaction terms and update the nodes recurrently can be described as \emph{message passing}
\shortcite{gilmer2017neural}, which propagates information across the graph.
In the gluing task, such learned information propagation may parallel the propagation of forces and other constraints over the structure.
For intuition, consider the tower in Figure~\ref{fig:task}. 
After one application of the edge model, the GN should be able to determine which block pairs are locally unstable, such as the top-most block in the figure, and thus require glue.
However, it does not have enough information to be able to determine that the bottom-most block in Figure~\ref{fig:task} also needs to be glued, because it is fully supporting the block above it.
Recurrent message-passing allows information about other blocks to be propagated to the bottom-most one, allowing for non-local relations to be reasoned about.

Given the updated edge, node, and global representations, we can decode them into edge-specific predictions, such as Q-values or unnormalized log probabilities (Figure~\ref{fig:graphnet}).
For the supervised setting, edges are glued with probability $p_{ij} \propto \mathrm{dec}_e(\mathbf{e}^\prime_{ij}; \theta_{\mathrm{dec}_e})$.
For the sequential decision making setting, we decode one action for each edge in the graph ($\pi_{ij}=\mathrm{dec}_e(\mathbf{e}_{ij}^\prime; \theta_{\mathrm{dec}_e})$) plus a ``stop'' action to end the gluing phase ($\pi_\sigma=\mathrm{dec}_g(\mathbf{g}^\prime; \theta_{\mathrm{dec}_g})$).

\begin{figure}[!t]
\begin{center}
\includegraphics[width=0.46\textwidth]{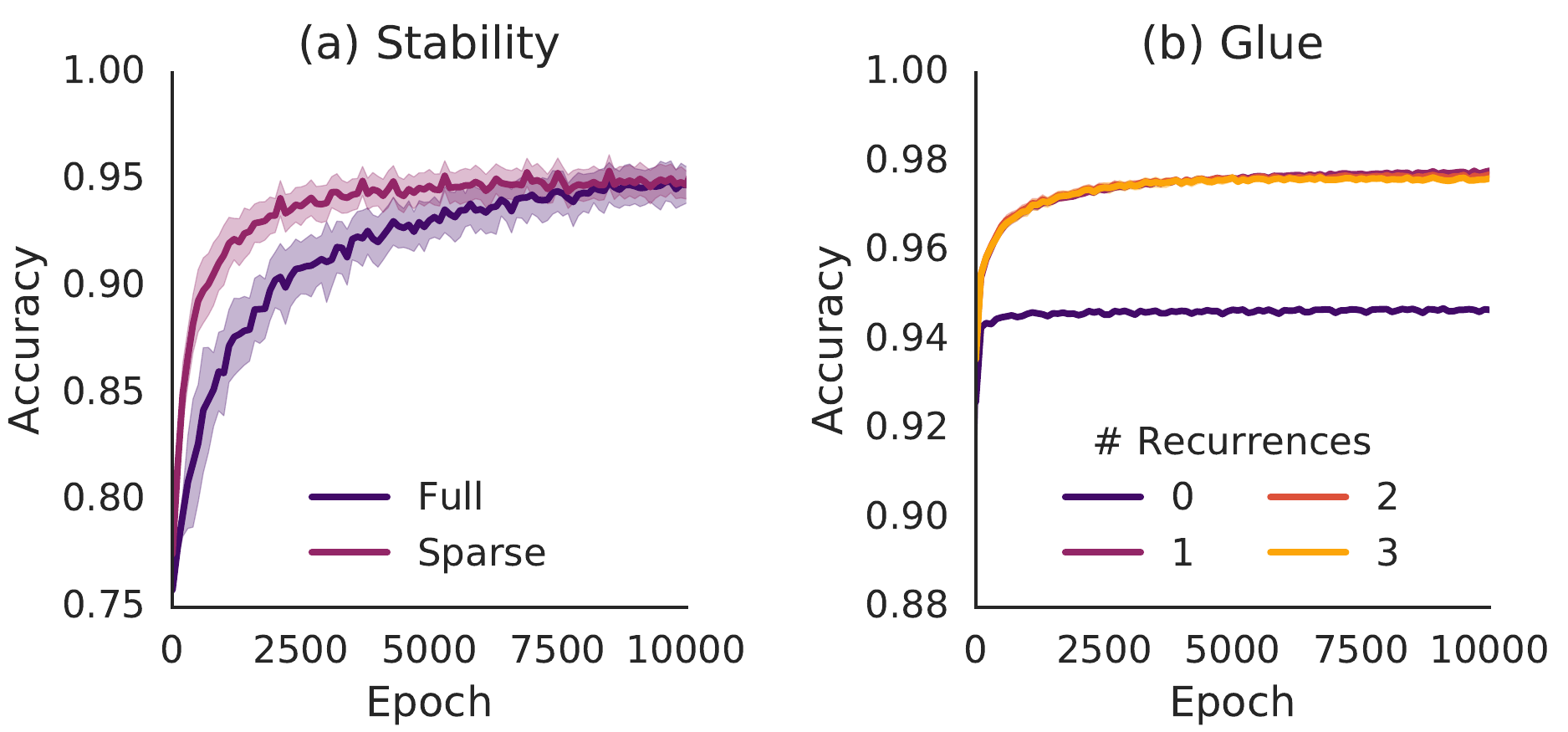}
\caption{\textbf{Supervised results for scenes with five blocks.} (a) Stability prediction for input graphs with contact information (sparse) or without (full). (b) Optimal glue prediction for models with different numbers of recurrent steps.}
\label{fig:stability-and-policy}
\end{center}
\end{figure}

\subsection{Supervised Learning Experiments}

Before investigating the full gluing task, we first explored how components of the graph network agent could perform key sub-tasks in a supervised setting, such as predicting stability or inferring which edges should be glued.

To test the GN's stability predictions, we used towers with variable number of blocks, where the input edges were labeled to indicate whether or not glue was present ($\mathbf{1}$ for glue, $\mathbf{0}$ for no glue). Glue was sampled randomly for each scene, and stability was defined as no blocks falling.
We tested two settings: \emph{fully connected} graphs (where the graph included all block-to-block edges), and \emph{sparse} graphs (where edges were only present between blocks that were in contact).
In both cases, GNs learned to accurately predict the stability of partially glued towers, but the sparse graph inputs yielded more efficient learning (Figure~\ref{fig:stability-and-policy}a). Results are shown for the case of 5 blocks, but these results are also consistent across towers with 6-9 blocks.
We also tested whether GNs can learn whether a contact between two blocks should be glued.
As discussed previously, some glue locations require reasoning about how forces propagate throughout the structure.
We therefore hypothesized that multiple message passing steps would be necessary to propagate this information, and indeed, we found that one recurrence was enough to dramatically improve glue prediction accuracy (Figure~\ref{fig:stability-and-policy}b).

\subsection{Sequential Decision Making Experiments}

From the supervised learning experiments, we concluded that GNs can accurately predict stability and select individual glue points.
Next we integrated these components into a full RL agent that performs the same gluing task that people faced, involving multiple actions and delayed rewards.

\subsubsection{Design}

We considered three agents: the \emph{multilayer perceptron} (or MLP) agent, the \emph{fully-connected graph network} (or GN-FC) agent, the \emph{graph network} (or GN) agent, and the \emph{simulation} agent.\footnote{Additional details about the agent architectures and training regimes are available in the appendix.}
As most deep RL agents are implemented either as MLPs or CNNs with no relational structure, our first agent chose actions according to a Q-function approximated by a MLP; as MLPs have a fixed input and output size, we trained a separate MLP for each tower size.
The GN and GN-FC agents (which had relational knowledge, but no explicit physical knowledge) also chose actions according to a Q-function and used 3 recurrent steps.
The GN agent used a sparse graph structure with edges corresponding to the contact points between the blocks, while the GN-FC used a fully connected graph structure and thus had to learn which edges corresponded to valid actions.
Finally, the simulation agent (which had both relational and physical knowledge) chose actions using simulation.
Specifically, for each unglued contact point, the agent ran a simulation to compute how many blocks would fall if that point were glued, and then chose the point which resulted in the fewest blocks falling.
This procedure was repeated until no blocks fell.
Note that the simulation agent is non-optimal as it chooses glue points greedily.

\subsubsection{The effect of relational structure}

\begin{figure}[t!]
\begin{center}
\includegraphics[width=0.46\textwidth]{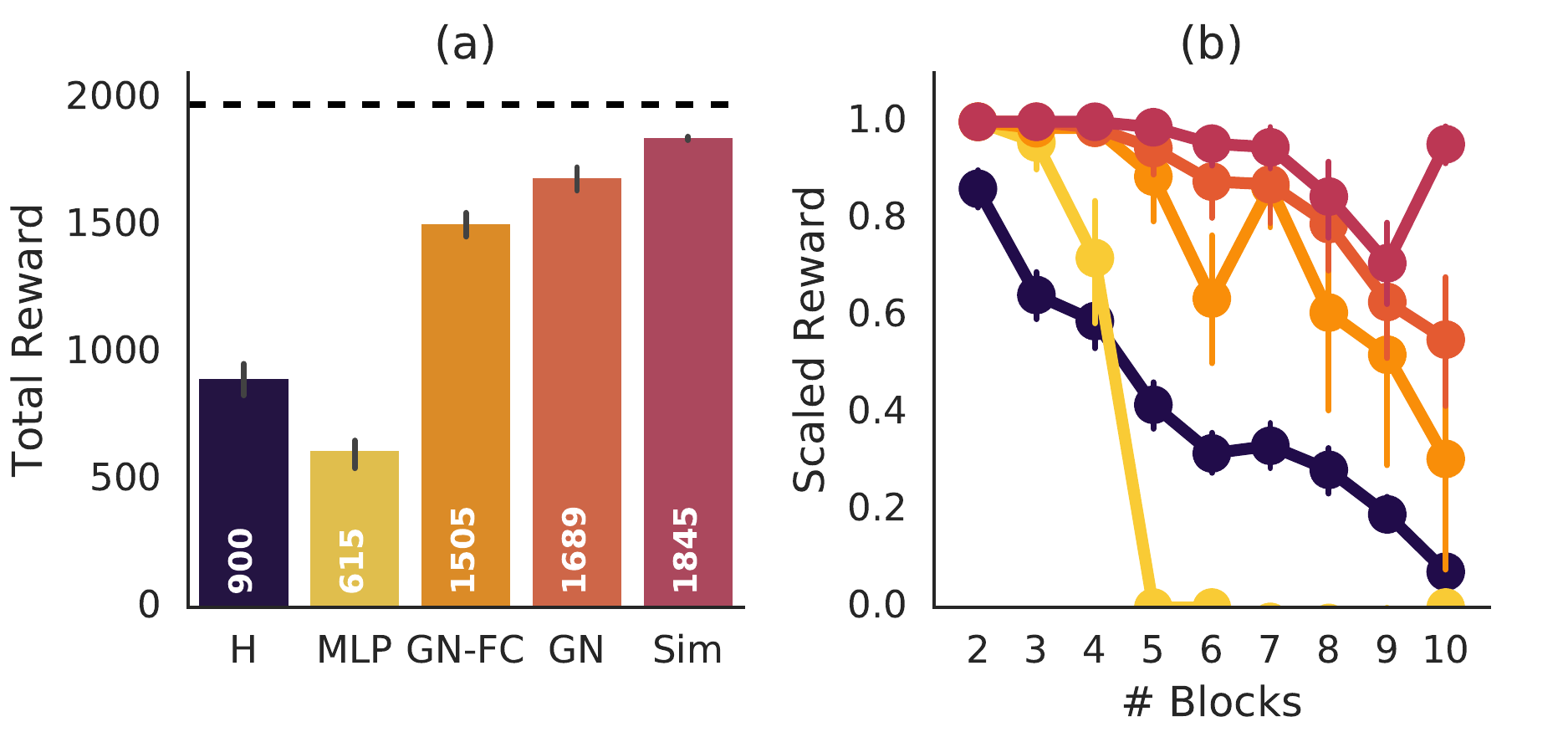}
\caption{(a) Comparison of overall reward for humans and agents. \emph{H}: human; \emph{MLP}: MLP agent; \emph{GN-FC}: GN agent operating over a fully-connected graph; \emph{GN}: GN agent operating over a sparse graph;
\emph{Sim}: simulation agent. (b) Comparison of scaled reward across towers of different sizes. Rewards are scaled such that 0 corresponds to the reward obtained when no actions are taken, and 1 to the optimal reward.
}
\label{fig:results}
\end{center}
\end{figure}

Both the MLP and the GN-FC agents take actions on the fully-connected graph (i.e., they both can choose pairs of blocks which are not adjacent); the main difference between them is that the GN-FC agent has a relational inductive bias while the MLP does not.
This relational inductive bias makes a large difference, with the GN-FC agent earning \GNFCVsMLPReward{} more points on average (Figure~\ref{fig:results}a) and also achieving more points across different tower sizes (Figure~\ref{fig:results}b).

Giving the correct relational structure in the GN agent further improves performance, with the GN agent achieving \GNVsGNFCReward{} more points on average than the GN-FC agent.
Thus, although the GN-FC agent does make use of relations, it does not always utilize the correct structure which ends up hurting its performance.
Indeed, we can observe that the GN-FC agent attempts invalid glue actions---for example, choosing edges between objects that are not adjacent, or self-edges---a whopping \GNFCPctInvalid{} of the time.
The MLP agent similarly picks ``invalid'' edges \MLPPctInvalid{} of the time.

The GN agents also exhibit much stronger generalization than the MLP agent.
To test generalization, we trained a second set of agents which did not observe towers of 7 or 10 blocks during training, and compared their test performance to our original set of agents.
The GN agent exhibited no detectable degradation in performance for either tower size, with a difference in scaled reward of \FullVsPartialSevenBlocks{} on 7-block towers and \FullVsPartialTenBlocks{} on 10-block towers.
The GN-FC agent interpolated successfully to 7-block towers (\FullVsPartialSevenBlocksFC{}), but struggled when extrapolating to 10-block towers (\FullVsPartialTenBlocksFC{}).
By definition, the MLP agent cannot generalize to new tower sizes because it is trained on each size independently.
We attempted to test for generalization anyway by training a single MLP on all towers and using zero-padding in the inputs for smaller towers.
However, this version of the MLP agent was unable to solve the task at all, achieving an average of \PaddedMLPReward{} points \emph{total}.

\subsubsection{The effect of physical knowledge}

The simulation agent was the only agent which incorporated explicit physical knowledge through its simulations, and we found that it also performed the best out of all the agents.
Specifically, the simulation agent earned on average \SimulationVsGNSReward{} points more than the GN agent, perhaps  suggesting that there is a benefit to using a model-based policy rather than a model-free policy (note, however, that the simulation agent has access to a perfect simulator; a more realistic implementation would likely fare somewhat worse).
However, we emphasize that the gain in performance by between the GN agent and the simulation agent was much less than that between the MLP and GN-FC agents, suggesting that relational knowledge may be more important than explicit physical knowledge in solving complex physical reasoning problems like the gluing task.

\subsubsection{Comparison to humans}

Although our goal was not to build a model of human cognition on the gluing task, we still compared people's behavior to that of the GN agent to elucidate any obvious differences.
Participants' average reward fell between the MLP and GN-FC agents' (Figure~\ref{fig:results}a).
As in Figure~\ref{fig:results}b, both agents and humans had increasing difficulty solving the task as a function of tower size, though this was expected: as the number of blocks in the tower increases, there is an exponential increase in the number of possible glue combinations.
Specifically, for a tower with $k$ contact points, there are $2^k$ possible ways glue can be applied (around 1000 possibilities for a 10-block tower), and optimally solving the task would require enumerating each of these possibilities.
Our agents do not do this, and it is unlikely that humans do either; therefore, the drop in performance as a function of tower size is not surprising.

Looking more closely, we found the GN agent made different patterns of errors than humans within scenes.
For example, while we found that people were more likely to make false positives (applying glue when none was needed), we did not find this to be true of the GN agent (\ModelProportionFP{} of errors, \ModelProportionFPTest{}).
This difference might be a result of perceptual uncertainty in humans, which leads to a tendency to over-estimate the instability of towers \shortcite{Battaglia2013}.

\section{Discussion}

In this paper, we explored the importance of relational inductive bias in performing interactive physical reasoning tasks.
We introduced a novel construction problem---the ``gluing task''---which involved gluing pairs of blocks together to stabilize a tower of blocks.
Our analysis showed that humans could perform far above chance and discovered they used systematic strategies, such as working top-to-bottom and reasoning about the whole glue configuration, before taking their first action.
Drawing on the view from cognitive psychology that humans understand the world in terms of objects and relations \shortcite{Shepard1980,Spelke2007,Kemp2008}, we developed a new deep RL agent that uses a decision-making policy based on object- and relation-centric representations, and measured its ability to learn to perform the gluing task.
These structured representations were instantiated using \emph{graph networks} (GNs), a family of neural network models that can be trained to approximate functions on graphs.
Our experiments showed that an agent with an object- and relation-centric policy could solve the task even better than humans, while an agent without such a \emph{relational inductive bias} performed far worse.
This suggests that a bias for acquiring relational knowledge is a key component of physical interaction, and can be effective even without an explicit model of physical dynamics.

Of course, model-based decision-making systems are powerful tools \shortcite{Silver2016}, and cognitive psychology work has found evidence that humans use internal physics models for physical prediction \shortcite{Battaglia2013}, inference \shortcite{Hamrick2016}, causal perception \shortcite{Gerstenberg2012}, and motor control \shortcite{Kawato1999}. Indeed, we found that the best performing agent in our task was the ``simulation'' agent, which used both relational and physical knowledge.
Provisioning deep RL agents with joint model-free and model-based strategies inspired by cognitive psychology has proven fruitful in imagination-based decision-making \shortcite{Hamrick2017}, and implementing relational inductive biases in similar systems should afford greater combinatorial generalization over state and action spaces.

More generally, the relational inductive bias possessed by our GN agent is not specific to physical scenes.
Indeed, certain aspects of human cognition have previously been studied and modeled in ways that are explicitly relational, such as in analogical reasoning \shortcite<e.g.>{gentner1983structure,Holyoak2012}.
In other cognitive domains, GNs might help capture how people build cognitive maps of their environments and use them to navigate; how they schedule their day to avoid missing important meetings; or how they decide whom to interact with at a cocktail party.
Each of these examples involves a set of entities, locations, or events which participate in interactive relationships and require arbitrarily complex relational reasoning to perform successfully.

In sum, this work demonstrates how deep RL can be improved by adopting relational inductive biases like those in human cognition, and opens new doors for developing formal cognitive models of more complex, interactive human behaviors like physical scene construction and interaction.

\subsubsection{Acknowledgements}

We would like to thank Tobias Pfaff, Sam Ritter, and anonymous reviewers for helpful comments.

\section{Supplementary Material}

\subsection{Architectural Details}

\subsubsection{MLP Agent}

The MLP agent had three hidden layers with 256 units each and ReLU nonlinearities.
Inputs to the MLP agent consisted of the concatenated positions and orientations of all objects in the scene, as well as a one-hot vector of size $E_{fc}=N(N-1)/2$ indicating which objects had glue between them.
There were $E_{fc}+1$ outputs in the final layer: one for each pair of blocks plus the floor (including non-adjacent objects), and an additional output for the ``stop'' action.

\subsubsection{GN Agents}

Th GN-FC agent had the same inputs and outputs as the MLP agent.
The inputs to the GN agent also included the positions and orientations of all objects in the scene, but the ``glue'' vector instead had size $E_{sparse}\approx N$ (where $E_{sparse}$ is the number of pairs of blocks in contact); the GN agent was also told \emph{which} blocks, specifically, were in contact.
There were $E_{sparse}+1$ outputs in the final layer.

Both GN agents used node, edge, and globals encoders which were each linear layers with an output dimensionality of size 64.
The edge, node, and global models were each a MLP with two hidden layers of 64 units (with ReLUs) and an output dimensionality of 64.
In these models we also used ``skip'' connections as in \shortciteA{Dai2017}, which means that we also fed in both encoded and non-encoded inputs to the model.
We additionally used a gated recurrent unit (GRU) as the core for our recurrent loop, similar to \shortciteA{LiTarlow2016}.
We passed the outputs of the recurrent GN to a second GN decoder (with the same architecture for the edge, node, and global models).
This second GN helps the agent decompose the problem, such as first detecting which block pairs are in contact, and then determining which of those pairs should be glued.
Finally, the edge and global values were further decoded by two hidden layers of 64 units (with ReLUs) and a final linear layer with a single output.

\subsection{Training Procedure}

Both the GN and MLP agents were trained for 300k episodes on 100k scenes for each tower size (900k total scenes), which were distinct from those in the behavioral experiment.
We used Q-learning with experience replay \shortcite{Mnih2015} with a replay ratio of 16, a learning rate of 1e-4, a batch size of 16, a discount factor of 0.9, and the Adam optimizer \shortcite{Kingma2015}.
Epsilon was annealed over 100k environment steps from 1.0 to 0.01.

Because the MLP agent had fixed input and output sizes that depend on the number of blocks in the scene, we trained nine separate MLP agents (one for each tower size).
Both GN agents were trained simultaneously on all tower sizes using a curriculum in which we began training on the next size tower (as well as all previous sizes) after every 10k episodes.

\bibliographystyle{apacite}
\setlength{\bibleftmargin}{.125in}
\setlength{\bibindent}{-\bibleftmargin}
\renewcommand\bibliographytypesize{\small}
\bibliography{CogSci_Template}

\end{document}